\documentclass[11pt,a4paper]{article}
\usepackage[hyperref]{naaclhlt2019}
\usepackage{times}
\usepackage{latexsym}
\usepackage{helvet}  %Required
\usepackage{courier}  %Required
\usepackage{url}  %Required
\usepackage{graphicx}  %Required
\graphicspath{{figures/}} % Directory in which figures are stored
\usepackage{amsmath}
\usepackage{amssymb}
\usepackage{xcolor} % for coloring fond
\usepackage{tabularx} % for automatically changing line in a cell of a table
\usepackage{multirow}
\usepackage{booktabs} % for toprule midrule and bottomrule in table
\usepackage{tablefootnote}
\usepackage[textsize=scriptsize]{todonotes}
\definecolor{blue}{RGB}{0, 93, 170}

\usepackage{soul}

\usepackage{url}

\aclfinalcopy % Uncomment this line for the final submission

\title{Multi-Granular Text Encoding for Self-Explaining Categorization}

\author{
  Zhiguo Wang$^1$, Yue Zhang$^2$, Mo Yu$^1$, Wei Zhang$^1$, Lin Pan$^1$\\ \textbf{Linfeng Song}$^3$, \textbf{Kun Xu}$^3$, \textbf{Yousef El-Kurdi}$^1$ \\
  $^1$IBM T.J. Watson Research Center, Yorktown Heights, NY 10598\\
  $^2$School of Engineering, Westlake University, China\\
  $^3$Tencent AI Lab\\
  {\tt zgw.tomorrow@gmail.com}\\
 }

\date{}

\begin{document}
\maketitle
\begin{abstract}
Self-explaining text categorization requires a classifier to
make a prediction along with supporting evidence.
A popular type of evidence is sub-sequences extracted from the input text which are sufficient for the classifier to make the prediction.
In this work, we define multi-granular ngrams as basic units for explanation, and organize all ngrams into a hierarchical structure, so that shorter ngrams can be reused while computing longer ngrams.
We leverage a tree-structured LSTM to learn a context-independent representation for each unit via parameter sharing.
Experiments on medical disease classification show that our model is more accurate, efficient and compact than BiLSTM and CNN baselines.
More importantly, our model can extract intuitive multi-granular evidence to support its predictions.
%Further experiments show that our model is sufficiently powerful to be used as a general text classification model, which outperforms the state-of-the-art models over several standard benchmarks.
\end{abstract}

\section{Introduction}
Increasingly complex neural networks have achieved highly competitive results for many NLP tasks \cite{vaswani2017attention,devlin2018bert}, 
but they prevent human experts from understanding
how and why a prediction is made. 
Understanding how a prediction is made can be very important for certain domains, such as the medical domain.
Recent research has started to investigate models with self-explaining capability, i.e. extracting evidence to support their final predictions \cite{li2015visualizing,lei2016rationalizing,lin2017structured,mullenbach2018explainable}.
For example, in order to make diagnoses based on the medical report in Table \ref{tab:example}, the highlighted symptoms may be extracted as evidence.

%Wei: Change this! trying to say "how highlighted words are used for classification"? Please cut to the chase.
Two methods have been proposed on \emph{how to jointly provide highlights along with classification}. 
(1) an \emph{extraction-based} method \citep{lei2016rationalizing}, which first extracts evidences from the original text and then makes a prediction solely based on the extracted evidences; 
(2) an \emph{attention-based} method \citep{lin2017structured,mullenbach2018explainable}, which leverages the self-attention mechanism to show the importance of basic units (words or ngrams) through their attention weights.

\begin{table}[tbp] \small
\small
\begin{tabularx}{210pt}{|X|}
\hline
\textbf{Medical Report:} The patient was admitted to the Neurological Intensive Care Unit for close observation. She was begun on \colorbox{red!40}{heparin anticoagulated} carefully secondary to the \colorbox{red!40}{petechial bleed}. She started weaning from the vent the next day.  She was started on Digoxin to control her rate and her Cardizem was held. She was started on antibiotics for possible \colorbox{red!40}{aspiration pneumonia}.  Her chest x-ray showed \colorbox{red!40}{retrocardiac effusion}.   She had some \colorbox{red!40}{bleeding after nasogastric tube insertion}. \\
\hline
\textbf{Diagnoses:} Cerebral artery occlusion;
Unspecified essential hypertension;
Atrial fibrillation;
Diabetes mellitus. \\
\hline
\end{tabularx}
%\vspace{-1.0em}
\caption{A medical report snippet and its diagnoses.}
\label{tab:example}
\vspace{-2.0em}
\end{table}

%Wei: Change this! trying to say "what the granularity and weights are for meaningful units to highlight"? Please cut to the chase.

However, previous work has several limitations.
\citet{lin2017structured}, for example, take single words as basic units, while meaningful information is usually carried by multi-word phrases. 
For instance, useful symptoms in Table \ref{tab:example}, such as ``bleeding after nasogastric tube insertion'', are larger than a single word.
Another issue of \citet{lin2017structured} is that their attention model  is applied on the representation vectors produced by an LSTM.
%Therefore, the highlighting in attention models is usually applied to a word's contextual encoding. For example, \citet{lin2017structured} uses LSTMs for context encoding.
Each LSTM output contains more than just the information of that position, thus the real range for the highlighted position is unclear.
\citet{mullenbach2018explainable} defines all 4-grams of the input text as basic units and uses a convolutional layer to learn their representations, which still suffers from fixed-length highlighting. Thus the explainability of the model is limited.
% \st{However, using isolate word positions as basic units of explainability can be rather weak.}\linfeng{, yet failed to capture these more meaningful phrasal evidences.} 
\citet{lei2016rationalizing} introduce a regularizer over the selected (single-word) positions to encourage the model to extract larger phrases.
However, their method can not tell how much a selected unit contributes to the model's decision through a weight value.
% \mo{also in order to avoid the information flow, they rely on RL which is more difficult to train and much slower}
% \citet{mullenbach2018explainable} define all 4-grams of the input text as basic units, and leverage a convolutional layer to learn representations for each 4-gram\linfeng{, yet natural phrases can have other than 4 words}.
% \st{However, constraining basic units to be fixed-length ngrams would limit the explainability of the model.} 

%Wei: " so that ... ' -> ' so as to flexibly highlight phrases as evidence, which previous approaches do not directly support.
In this paper, we study \emph{what the meaningful units to highlight are}.
%To address these limitations, 
We define multi-granular ngrams as basic units, so that all highlighted symptoms in Table \ref{tab:example} can be directly used for explaining the model.
%Our model borrows the idea from tree-structured LSTMs \cite{tai2015improved,zhu2015long,teng2016bidirectional}.
Different ngrams can have overlap.
To improve the efficiency, we organize all ngrams into a hierarchical structure, such that the shorter ngram representations can be reused to construct longer ngram representations.
Experiments on medical disease classification show that our model is more accurate, efficient and compact than BiLSTM and CNN baselines.
Furthermore, our model can extract intuitive multi-granular evidence to support its predictions.
%Additionally, we show that our model could serve as a powerful text encoding component for general text categorization tasks. %, where our model outperforms state-of-the-art models on several standard benchmarks. 
%\linfeng{About section 2: It's not very clear on the specific relation between our method and the two baselines. My understanding is that \emph{Text Encoder} can only be BiLSTM, and the layer above is our contribution. It may not be clear to reviewers.}

\begin{figure}[tbp]
\begin{center}
\includegraphics[width=0.35\textwidth]{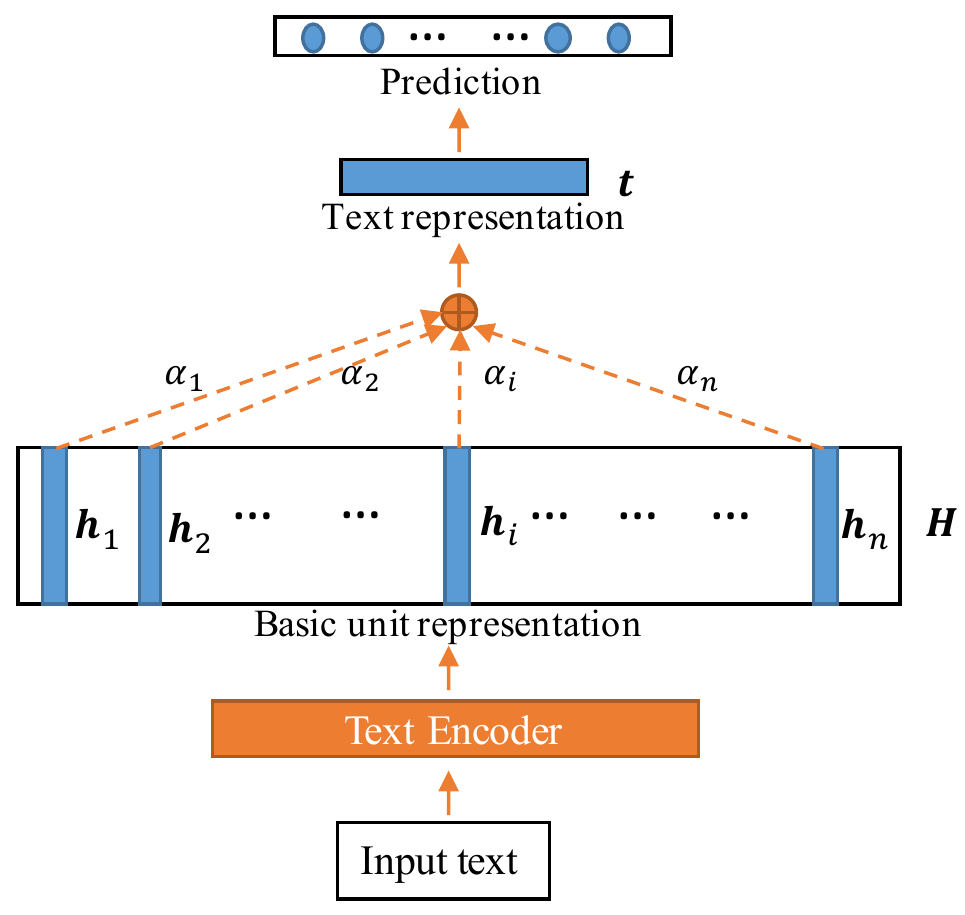}
\end{center}
\vspace{-1.0em}
\caption{A generic architecture.}
\label{fig:generic_architecture}
\vspace{-1.5em}
\end{figure}

\section{Generic architecture and baselines}
%We aim to design a text categorization model, which can pinpoint evidences to support its categorization predictions.
Our work leverages the \emph{attention-based} self-explaining method \citep{lin2017structured}, as shown in Figure \ref{fig:generic_architecture}.
First, our text encoder (\S \ref{sec:main_method}) formulates an input text into a list of basic units, 
learning a vector representation for each,
where the basic units can be words, phrases, or arbitrary ngrams.
Then, the attention mechanism is leveraged over all basic units, 
and sums up all unit representations based on the attention weights \{$\alpha_1,..., \alpha_n$\}. 
Eventually, the attention weight $\alpha_i$ will be used to reveal how important a basic unit $\boldsymbol{h}_i$ is.
The last prediction layer takes the fixed-length text representation $\boldsymbol{t}$ as input, and makes the final prediction.

\textbf{Baselines}: We compare two types of baseline text encoders in Figure \ref{fig:generic_architecture}. 
(1) \textbf{\citet{lin2017structured} (BiLSTM)}, which formulates single word positions as basic units, and computes the vector $\boldsymbol{h}_i$ for the $i$-th word position with a BiLSTM;
(2) \textbf{Extension of \citet{mullenbach2018explainable} (CNN)}. The original model in \cite{mullenbach2018explainable} only utilizes 4-grams. Here we extend this model to take all unigrams, bigrams, and up to $n$-grams as the basic units.

For a fair comparison, both our approach and the baselines share the same architecture, and the only difference is the text encoder used.

\begin{figure}[tbp]
\begin{center}
\includegraphics[width=0.35\textwidth]{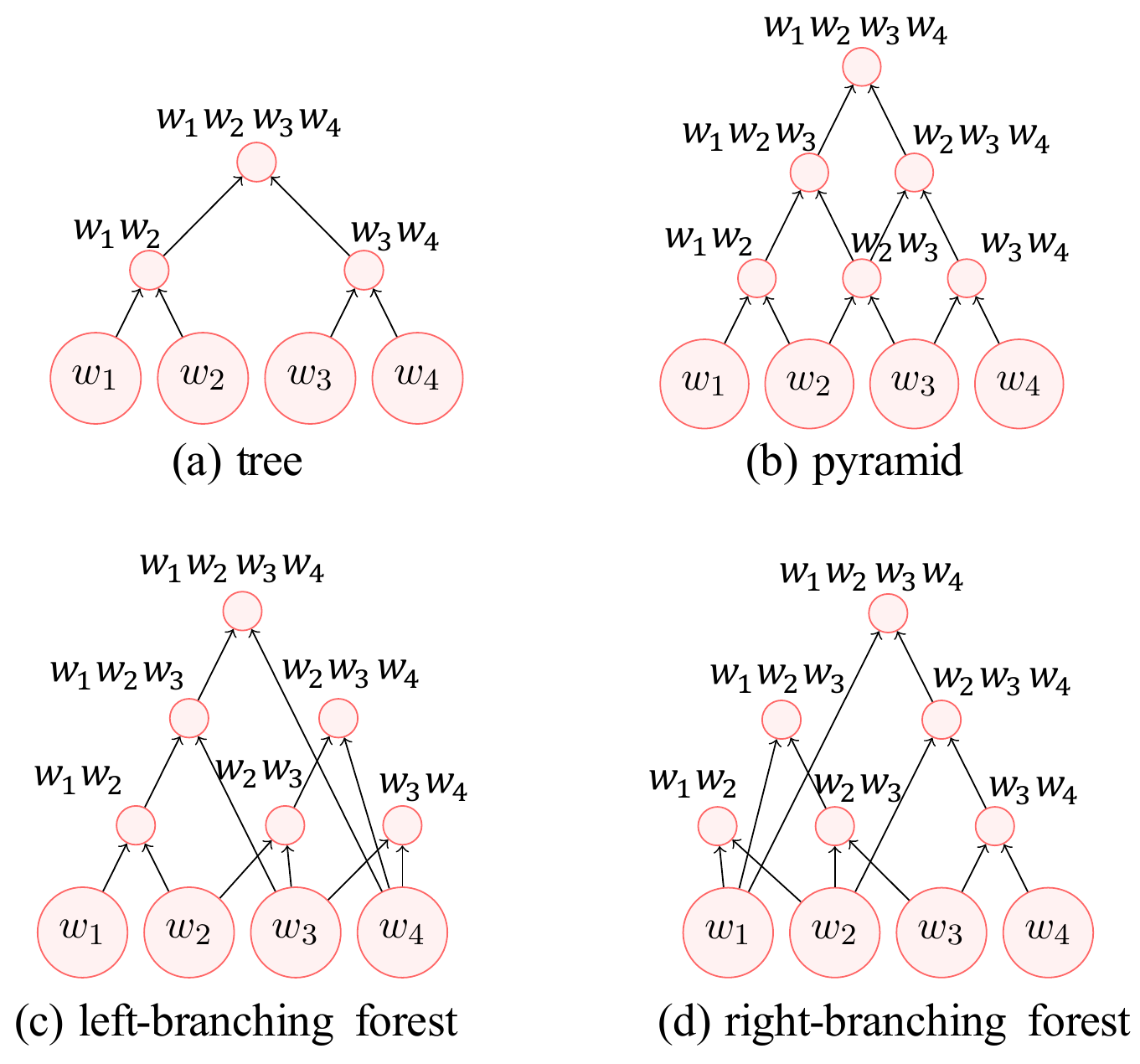}
\end{center}
\vspace{-1.0em}
\caption{Structures for a sentence $w_1w_2w_3w_4$, where each node corresponds to a phrase or ngram.}
\label{fig:diagram}
\vspace{-1.5em}
\end{figure}

%\textbf{Limitation of baselines:} \linfeng{This part has been mentioned in the introduction, so maybe we can remove it for more space?}
%For the BiLSTM baseline, the model can only pinpoint important words to support its predictions, and the representation of each word position encodes information from its sentence-level contexts, 
%which makes it difficult to tell whether an higher attention score is from the word itself or its context. 
%The CNN baseline does not reuse representations of shorter ngrams while computing longer ngrams, therefore it neglects the intrinsic relationship between ngrams and the computation process is inefficient (compared to our proposed approach).
%Besides, it has to define different filters for different order of ngrams, so it requires much more parameters than the BiLSTM baseline.

\section{Multi-granular text encoder}
\label{sec:main_method}

We propose the multi-granular text encoder to deal with drawbacks (as mentioned in the third paragraph of Section 1) of our baselines.

\textbf{Structural basic units}: 
We define basic units for the input text as multi-granular ngrams, organizing ngrams in four different ways.
Taking a synthetic sentence $w_1w_2w_3w_4$ as the running example, 
we illustrate these structures in Figure \ref{fig:diagram} (a), (b), (c) and (d), respectively.
The first is a tree structure (as shown in Figure \ref{fig:diagram}(a)) that includes all phrases from a (binarized) constituent tree over the input text, 
where no cross-boundary phrases exists. 
The second type (as shown in Figure \ref{fig:diagram} (b,c,d)) includes all possible ngrams from the input text,  which is a superset of the tree structure.
In order to reuse representations of smaller ngrams while encoding bigger ngrams, 
all ngrams are organized into hierarchical structures in three different ways.
First, inspired by \citet{zhao2015self}, a pyramid structure is created for all ngrams as shown in
Figure \ref{fig:diagram}(b),
where leaf nodes are unigrams of the input text, and higher level nodes correspond to higher-order ngrams.
A disadvantage of the pyramid structure is that some lower level nodes may be used repeatedly while encoding higher level nodes, which may improperly augment the influence of the repeated units. 
For example, when encoding the trigram node ``$w_1w_2w_3$'', the unigram node ``$w_2$'' is used twice through two bigram nodes ``$w_1w_2$'' and ``$w_2w_3$''.
To tackle this issue, a left-branching forest structure is constructed for all ngrams as shown in Figure \ref{fig:diagram}(c),
where ngrams with the same prefix are grouped together into a left-branching binary tree, and, in this arrangement, multiple trees construct a forest.
Similarly, we construct a right-branching forest as shown in Figure \ref{fig:diagram}(d).

\textbf{Encoding}:  
We leverage a tree-structured LSTM composition function \cite{tai2015improved,zhu2015long,teng2016bidirectional} to compute node embeddings for all structures in Figure \ref{fig:diagram}.
Formally, the \textbf{state} of each node is represented as a pair of one hidden vector $\boldsymbol{h}$ and one memory representation $\boldsymbol{c}$,
which are calculated by composing the node's label embedding $\boldsymbol{x}$ and states of its left child $\langle \boldsymbol{h}^{\,l}, \boldsymbol{c}^{\,l}\rangle$ and right child $\langle \boldsymbol{h}^{\,r}, \boldsymbol{c}^{\,r}\rangle$ with gated functions:
\begin{align}
\boldsymbol{i} &=\sigma(W^1 \boldsymbol{x} + U_l^1 \boldsymbol{h}^{\,l} + U_r^1 \boldsymbol{h}^{\,r} + b^1) \\
\boldsymbol{f}^{\,l} &=\sigma(W^2 \boldsymbol{x} + U_l^2 \boldsymbol{h}^{\,l} + U_r^2 \boldsymbol{h}^{\,r} + b^2)\\
\boldsymbol{f}^{\,r} &=\sigma(W^3 \boldsymbol{x} + U_l^3 \boldsymbol{h}^{\,l} + U_r^3 \boldsymbol{h}^{\,r} + b^3)\\
\boldsymbol{o} &=\sigma(W^4 \boldsymbol{x} + U_l^4 \boldsymbol{h}^{\,l} + U_r^4 \boldsymbol{h}^{\,r} + b^4)\\
\boldsymbol{u} &=\tanh(W^5 \boldsymbol{x} + U_l^5 \boldsymbol{h}^{\,l} + U_r^5 \boldsymbol{h}^{\,r} + b^5)\\
\boldsymbol{c} &= \boldsymbol{i} \odot \boldsymbol{u} + \boldsymbol{f}^{\,l} \odot \boldsymbol{h}^{\,l} + \boldsymbol{f}^{\,r} \odot \boldsymbol{h}^{\,r}\\
\boldsymbol{h} &=  \boldsymbol{o} \odot \tanh(\boldsymbol{c})
\end{align}
where $\sigma$ is the sigmoid activation function, $\odot$ is the elementwise product,
$\boldsymbol{i}$ is the input gate, $\boldsymbol{f}^{\,l}$ and $\boldsymbol{f}^{\,r}$ are the forget gates for the left and right child respectively, and $\boldsymbol{o}$ is the output gate. 
We set $\boldsymbol{x}$ as the pre-trained word embedding for leaf nodes, and zero vectors for other nodes.
The representations for all units (nodes) can be obtained by encoding each basic unit in a bottom-up order.

\textbf{Comparison with baselines}
%Similar to the CNN baseline, 
%our text encoder defines multi-granular ngrams as basic units, 
%and separates the influence of contextual information from a unit's own representation.
%If more contexts are required, we only need to increase the depth in order to choose basic units covering more words.
Our encoder is more efficient than CNN while encoding bigger ngrams, because it reuses representations of smaller ngrams.
Furthermore, the same parameters are shared across all ngrams, which makes our encoder more compact, 
whereas the CNN baseline has to define different filters for different order of ngrams, so it requires much more parameters.
%Similar to the BiLSTM baseline, our text encoder utilizes the gated mechanism to encode ngrams.
Experiments show that using basic units 
up to 7-grams to construct the forest structure is good enough, which makes our encoder more efficient than BiLSTM.
Since in our encoder, all ngrams with the same order can be computed in parallel, 
and the model needs at most 7 iterative steps along the depth dimension for representing a given text of arbitrary length.

\section{Experiments}

\textbf{Dataset}:
We experiment on a public medical text classification dataset.\footnote{https://github.com/SnehaVM/Medical-Text-Classification} 
Each instance consists of a medical abstract with an average length of 207 tokens, and one label out of five categories indicating which disease this document is about. 
We randomly split the dataset into train/dev/test sets by 8:1:1 for each category, 
and end up with 11,216/1,442/1,444 instances for each set.

\textbf{Hyperparameters}
We use the 300-dimensional GloVe word vectors pre-trained from the 840B Common Crawl corpus \cite{pennington2014glove}, and  set the hidden size as 100 for node embeddings.
We apply dropout to every layer with a dropout ratio 0.2, and set the batch size as 50.
We minimize the cross-entropy of the training set with the ADAM optimizer \cite{kingma2014adam}, and set the learning rate is to 0.001. 
During training, the pre-trained word embeddings are not updated. 

\begin{figure}[tbp]
\begin{center}
\includegraphics[width=0.4\textwidth]{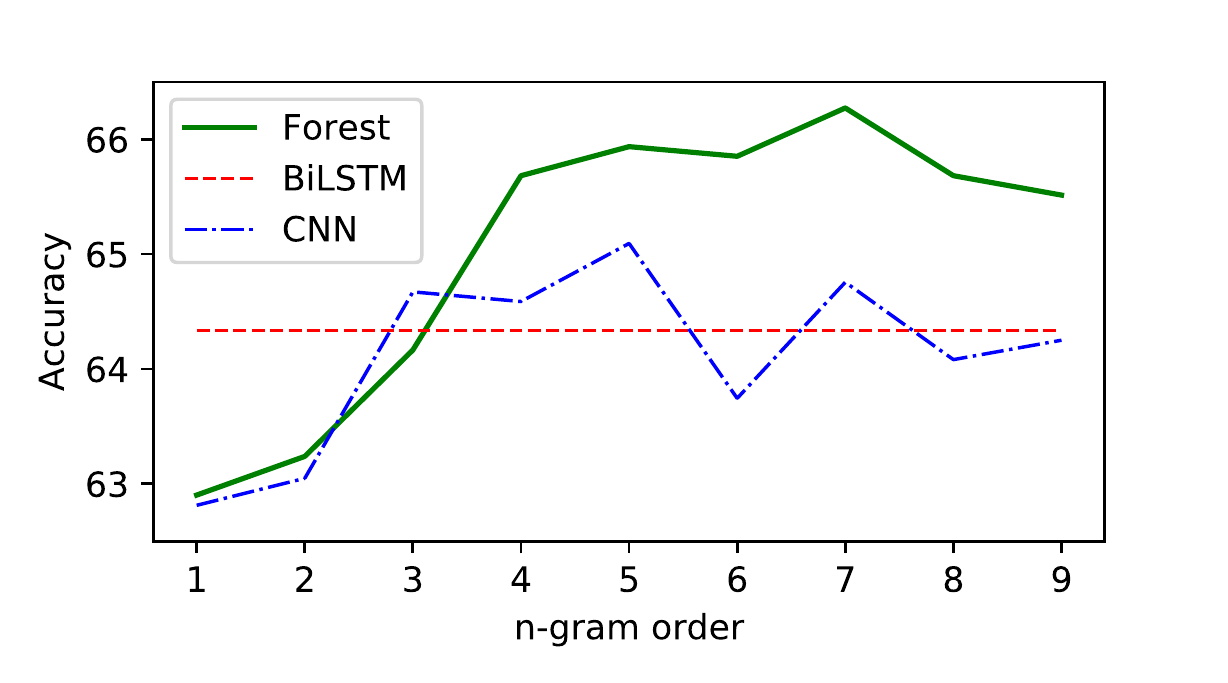}
\end{center}
\vspace{-1.0em}
\caption{Influence of n-gram order.}
\label{fig:ngram_order}
\vspace{-0.5em}
\end{figure}

\subsection{Properties of the multi-granular encoder}
\textbf{Influence of the n-gram order}: 
%We choose the left-branching forest structure in Figure \ref{fig:diagram}(c) to structuralize ngrams.
For CNN and our \emph{LeftForest} encoder, we vary the order of ngrams from 1 to 9,
and plot results in Figure \ref{fig:ngram_order}.
For BiLSTM,  we draw a horizontal line according to its performance, since the ngram order does not apply.
When ngram order is less than 3, both CNN and \emph{LeftForest} underperform BiLSTM.
When ngram order is over 3, \emph{LeftForest} outperforms both baselines.
%and the curve saturates after ngram order reaches 7.
Therefore, in terms of accuracy, our multi-granular text encoder is more powerful than baselines.

\begin{table}[tbp]
\small
\begin{center}
\begin{tabular}{lcccc}
\toprule
Model & Train Time & Eval Time & ACC & \#Param.\\
\midrule
CNN & 57.0 & 2.6 & 64.8 & 848,228 \\
BiLSTM & 92.1 & 4.6 & 64.5 & 147,928 \\
LeftForest & 30.3 & 1.4 & 66.2 & 168,228 \\
\bottomrule
\end{tabular}
\end{center}
\vspace{-0.5em}
\caption{Efficiency evaluation.}
\label{tab:effi_eval}
\vspace{-1.0em}
\end{table}

\textbf{Efficiency}:
We set ngram order as 7 for both CNN and our encoder.
Table \ref{tab:effi_eval} shows the time cost (seconds) of one iteration over the training set and evaluation on the development set.
BiLSTM is the slowest model, because it has to scan over the entire text sequentially.
\emph{LeftForest} is almost 2x faster than CNN, because \emph{LeftForest} reuses lower-order ngrams while computing higher-order ngrams.
This result reveals that our encoder is more efficient than baselines.

\textbf{Model size}: 
In Table \ref{tab:effi_eval}, the last two columns show the
accuracy and number of parameters for each model.
\emph{LeftForest} contains much less parameters than CNN, and it gives a better accuracy than BiLSTM with only a small amount of extra parameters.
Therefore, our encoder is more compact.

\subsection{Model performance}

\begin{table}[tbp]
\small
\begin{center}
\begin{tabular}{lc}
\toprule
Model & Accuracy\\
\toprule
BiLSTM	& 62.7 \\
CNN	& 62.5 \\
\midrule
Tree & 63.8 \\
Pyramid & 63.7 \\
LeftForest & 64.6 \\
RightForest & 64.5 \\
BiForest & \textbf{65.2} \\
\bottomrule
\end{tabular}
\end{center}
\vspace{-1.0em}
\caption{Test set results.}
\label{tab:classification_test}
\vspace{-0.5em}
\end{table}

Table \ref{tab:classification_test} lists the accuracy on the test set, where \emph{BiForest} represents each ngram by concatenating representations of this ngram from both the \emph{LeftForest} and the \emph{RightForest} encoders.
We get several interesting observations: 
(1) Our multi-granular text encoder outperforms both the CNN and BiLSTM baselines regardless of the structure used;
(2) The \emph{LeftForest} and \emph{RightForest} encoders work better than the \emph{Tree} encoder, 
which shows that representing texts with more ngrams is helpful than just using the non-overlapping phrases from a parse tree; 
(3) The \emph{LeftForest} and \emph{RightForest} encoders give better performance than the \emph{Pyramid} encoder, 
which verifies the advantages of organizing ngrams with forest structures;
(4) There is no significant difference between the \emph{LeftForest} encoder and the \emph{RightForest} encoder. However, by combining them, the \emph{BiForest} encoder gets the best performance among all models, indicating that the \emph{LeftForest} encoder and the \emph{RightForest} encoder complement each other for better accuracy.

\begin{figure}[tbp]
\begin{center}
\includegraphics[width=0.4\textwidth]{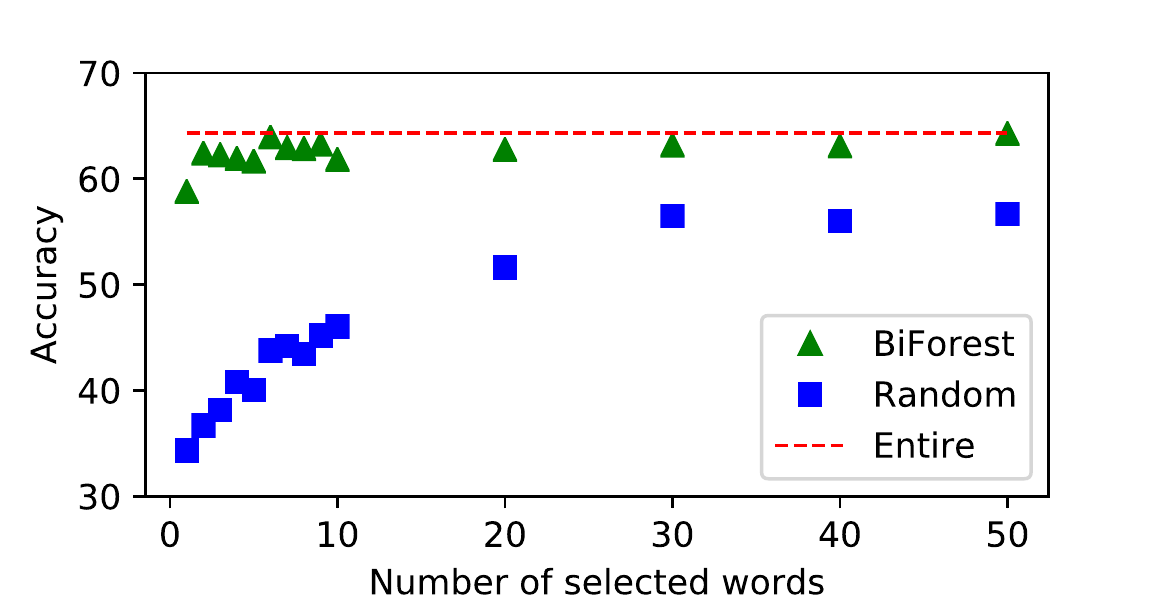}
\end{center}
\vspace{-1.0em}
\caption{Effectiveness of the extracted evidence.}
\label{fig:salient_evaluation}
\vspace{-1.5em}
\end{figure}

\subsection{Analysis of explainability}

\paragraph{Qualitative analysis} 
The following text is a snippet of an example from the dev set.
We leverage our \emph{BiForest} model to extract ngrams whose attention scores are higher than 0.05, and use the bold font to highlight them.
We extracted three ngrams as supporting evidence for its predicted category ``nervous system diseases''.
Both the \emph{spontaneous extradural spinal hematoma} and the \emph{spinal arteriovenous malformation} are diseases related to the spinal cord, 
therefore they are good evidence to indicate the text is about ``nervous system diseases''.

\textbf{\small Snippet}: \textit{
\small
Value of magnetic resonance imaging in \textbf{spontaneous extradural spinal hematoma} due to vascular malformation : case report . A case of spinal cord compression due to \textbf{spontaneous extradural spinal hematoma} is reported . A \textbf{spinal arteriovenous malformation} was suspected on the basis of magnetic resonance imaging. 
Early surgical exploration allowed a complete neurological recovery . 
%The vascular malformation was histopathologically confirmed . The role of magnetic resonance imaging in the evaluation of acute spinal cord compression syndromes is stressed .
}

\paragraph{Quantitative analysis} 
For each instance in the training set and the dev set, we 
utilize the attention scores from \emph{BiForest} to sort all ngrams, and create different copies of the training set and development set by only keeping the first $n$ important words.
We then train and evaluate a BiLSTM model with the newly created dataset.
We vary the number of words $n$ among \{1, 2, 3, 4, 5, 6, 7, 8, 9, 10, 20, 30, 40, 50\}, and show the corresponding accuracy with the green triangles in Figure \ref{fig:salient_evaluation}.
We define a \emph{Random} baseline by randomly selecting a sub-sequence containing $n$ words,
and plot its accuracy with blue squares in Figure \ref{fig:salient_evaluation}.
We also take a BiLSTM model trained with the entire texts as the upper bound (the horizontal line in Figure \ref{fig:salient_evaluation}).
When using only a single word for representing instances, single words extracted from our \emph{BiForest} model are significantly more effective than randomly picked single words.
When utilizing up to five extracted words from our \emph{BiForest} model for representing each instance, we can obtain an accuracy very close to the upper bound.
Therefore, the extracted evidence from our \emph{BiForest} model are truly effective for representing the instance and its corresponding category.

\section{Conclusion}
We proposed a multi-granular text encoder for self-explaining text categorization.
Comparing with the existing BiLSTM and CNN baselines, our model is more accurate, efficient and compact. In addition, our model can extract effective and intuitive evidence to support its predictions.

\bibliography{main}

\begin{thebibliography}{12}
\expandafter\ifx\csname natexlab\endcsname\relax\def\natexlab#1{#1}\fi

\bibitem[{Devlin et~al.(2018)Devlin, Chang, Lee, and
  Toutanova}]{devlin2018bert}
Jacob Devlin, Ming-Wei Chang, Kenton Lee, and Kristina Toutanova. 2018.
\newblock Bert: Pre-training of deep bidirectional transformers for language
  understanding.
\newblock \emph{arXiv preprint arXiv:1810.04805}.

\bibitem[{Kingma and Ba(2014)}]{kingma2014adam}
Diederik Kingma and Jimmy Ba. 2014.
\newblock Adam: A method for stochastic optimization.
\newblock \emph{arXiv preprint arXiv:1412.6980}.

\bibitem[{Lei et~al.(2016)Lei, Barzilay, and Jaakkola}]{lei2016rationalizing}
Tao Lei, Regina Barzilay, and Tommi Jaakkola. 2016.
\newblock Rationalizing neural predictions.
\newblock \emph{arXiv preprint arXiv:1606.04155}.

\bibitem[{Li et~al.(2015)Li, Chen, Hovy, and Jurafsky}]{li2015visualizing}
Jiwei Li, Xinlei Chen, Eduard Hovy, and Dan Jurafsky. 2015.
\newblock Visualizing and understanding neural models in nlp.
\newblock \emph{arXiv preprint arXiv:1506.01066}.

\bibitem[{Lin et~al.(2017)Lin, Feng, Santos, Yu, Xiang, Zhou, and
  Bengio}]{lin2017structured}
Zhouhan Lin, Minwei Feng, Cicero Nogueira~dos Santos, Mo~Yu, Bing Xiang, Bowen
  Zhou, and Yoshua Bengio. 2017.
\newblock A structured self-attentive sentence embedding.
\newblock \emph{arXiv preprint arXiv:1703.03130}.

\bibitem[{Mullenbach et~al.(2018)Mullenbach, Wiegreffe, Duke, Sun, and
  Eisenstein}]{mullenbach2018explainable}
James Mullenbach, Sarah Wiegreffe, Jon Duke, Jimeng Sun, and Jacob Eisenstein.
  2018.
\newblock Explainable prediction of medical codes from clinical text.
\newblock \emph{arXiv preprint arXiv:1802.05695}.

\bibitem[{Pennington et~al.(2014)Pennington, Socher, and
  Manning}]{pennington2014glove}
Jeffrey Pennington, Richard Socher, and Christopher~D Manning. 2014.
\newblock Glove: Global vectors for word representation.
\newblock In \emph{EMNLP}.

\bibitem[{Tai et~al.(2015)Tai, Socher, and Manning}]{tai2015improved}
Kai~Sheng Tai, Richard Socher, and Christopher~D Manning. 2015.
\newblock Improved semantic representations from tree-structured long
  short-term memory networks.
\newblock \emph{arXiv preprint arXiv:1503.00075}.

\bibitem[{Teng and Zhang(2016)}]{teng2016bidirectional}
Zhiyang Teng and Yue Zhang. 2016.
\newblock Bidirectional tree-structured lstm with head lexicalization.
\newblock \emph{arXiv preprint arXiv:1611.06788}.

\bibitem[{Vaswani et~al.(2017)Vaswani, Shazeer, Parmar, Uszkoreit, Jones,
  Gomez, Kaiser, and Polosukhin}]{vaswani2017attention}
Ashish Vaswani, Noam Shazeer, Niki Parmar, Jakob Uszkoreit, Llion Jones,
  Aidan~N Gomez, {\L}ukasz Kaiser, and Illia Polosukhin. 2017.
\newblock Attention is all you need.
\newblock In \emph{Advances in Neural Information Processing Systems}, pages
  6000--6010.

\bibitem[{Zhao et~al.(2015)Zhao, Lu, and Poupart}]{zhao2015self}
Han Zhao, Zhengdong Lu, and Pascal Poupart. 2015.
\newblock Self-adaptive hierarchical sentence model.
\newblock In \emph{IJCAI}, pages 4069--4076.

\bibitem[{Zhu et~al.(2015)Zhu, Sobihani, and Guo}]{zhu2015long}
Xiaodan Zhu, Parinaz Sobihani, and Hongyu Guo. 2015.
\newblock Long short-term memory over recursive structures.
\newblock In \emph{International Conference on Machine Learning}, pages
  1604--1612.

\end{thebibliography}
\bibliographystyle{acl_natbib}

\end{document}